\documentclass[conference]{IEEEtran}
\usepackage{times}

\usepackage[numbers]{natbib}
\usepackage{multicol}
\usepackage[bookmarks=true]{hyperref}
\usepackage{mwe}
\usepackage{floatrow}
\usepackage{framed}
\begin{document}

% paper title
\title{Multimodal Data Fusion for Power-On-and-Go Robotic Systems in Retail}

% You will get a Paper-ID when submitting a pdf file to the conference system
% \author{Author Names Omitted for Anonymous Review. Paper-ID [add your ID here]}
\author{\authorblockN{Shubham Sonawani}
\authorblockA{Interactive Robotics Lab\\ Arizona State University} 
\and 
\authorblockN{Kailas Maneparambil}
\authorblockA{Embedded Systems Lab\\ Intel}
\and
\authorblockN{Heni Ben Amor}
\authorblockA{Interactive Robotics Lab\\ Arizona State University}}
\maketitle

\begin{abstract}
Robotic systems for retail have gained a lot of attention due to the labor-intensive nature of such business environments. Many tasks have the potential to be automated via intelligent robotic systems that have manipulation capabilities. For example, empty shelves can be replenished, stray products can be picked up or new items can be delivered. However, many challenges make the realization of this vision a challenge. In particular, robots are still too expensive and do not work out of the box. In this paper, we discuss a work-in-progress approach for enabling power-on-and-go robots in retail environments through a combination of active, physical sensors and passive, artificial sensors. In particular, we use low-cost hardware sensors in conjunction with machine learning techniques in order to generate high-quality environmental information. More specifically, we present a setup in which a standard monocular camera and Bluetooth low-energy yield a reliable robot system that can immediately be used after placing a couple of sensors in the environment. The camera information is used to synthesize accurate 3D point clouds, whereas the BLE data is used to integrate the data into a complex map of the environment. The combination of active and passive sensing enables high-quality sensing capabilities at a fraction of the costs traditionally associated with such tasks. 

% Need more description
\end{abstract}
\IEEEpeerreviewmaketitle
\section{Introduction}
Robotics technology has made great strides towards autonomous systems in important application
domains such as surgery, flexible manufacturing, or space exploration.  In recent years, there is
increased interest in the application of such technologies in other, more conservative, domains that
have traditionally been averse to such innovations.  In particular, retail stores and warehouses \cite{lee2019robotics, corbato2018integrating} are at
the cusp of a robotic revolution. However, despite the obvious potential, a number of critical issues have hampered the actual adoption of mobile robot systems in such businesses. Chief among these issues and concerns are a.) acquisition costs and b.) the constant need for calibration and human intervention. Hence, there is substantial interest in efficient, low-cost robotic systems that can self-initialize and calibrate while also generating high-quality sensor data of the environment. These objectives, however, can seem at odds with each other.

In this paper, we present our ongoing work regarding ``power-on-and-go" (PO\&G) robot systems for retail and warehouses. The main rationale underlying our approach is that multimodal integration between physical sensors and artificial sensors enables cost-efficient, data-rich, and redundant sensing of the robot workspace. The expression ``artificial sensors" in this context refers to algorithms that can synthesize complex information from low-cost/low-quality data streams. Rather than relying on expensive (and typically heavy) sensor devices, i.e., LIDAR, we train machine learning (ML) models that can generate similar high-quality data from affordable sensors, i.e., a simple monocular camera. The trained "artificial sensors" not only reduce costs but also allow us to bypass current limitations of high-end sensors, e.g., limited image resolution.
\begin{figure}[t!]
    \centering
    \includegraphics[width=0.8\textwidth]{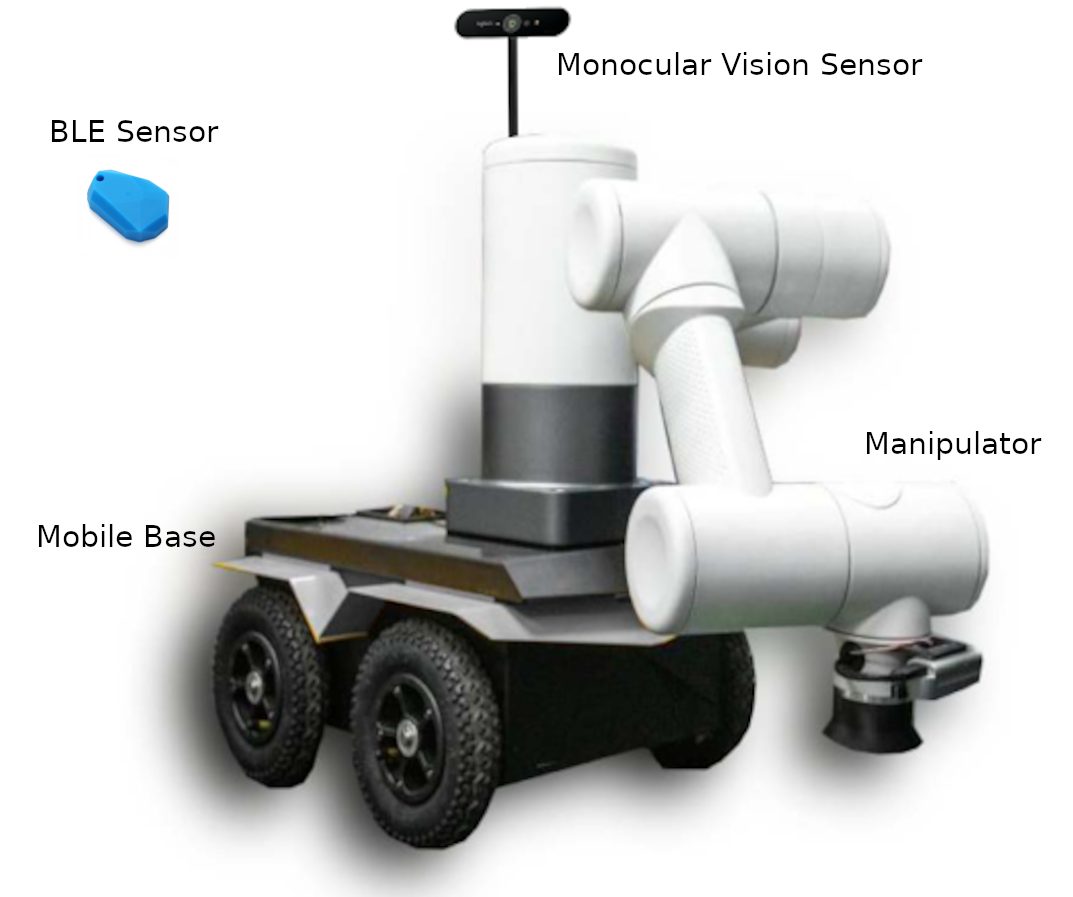}
    \caption{A power-on-and-go robotics system for retail. Depth estimation and BLE sensors are used to ensure reliable, affordable and high-quality sensing for navigation and manipulation.}
    \label{fig:robot}
\end{figure}
At the same time, such learned sensing models also come with potential pitfalls. Building models of the environment purely based on inferred data from learned models without a physical grounding can lead to accumulating errors. For example, repeated estimation of the location of objects in the environment can lead to compounding errors and substantial drift over time.

\begin{figure*}[t!]
    \centering
    \includegraphics[width=\textwidth]{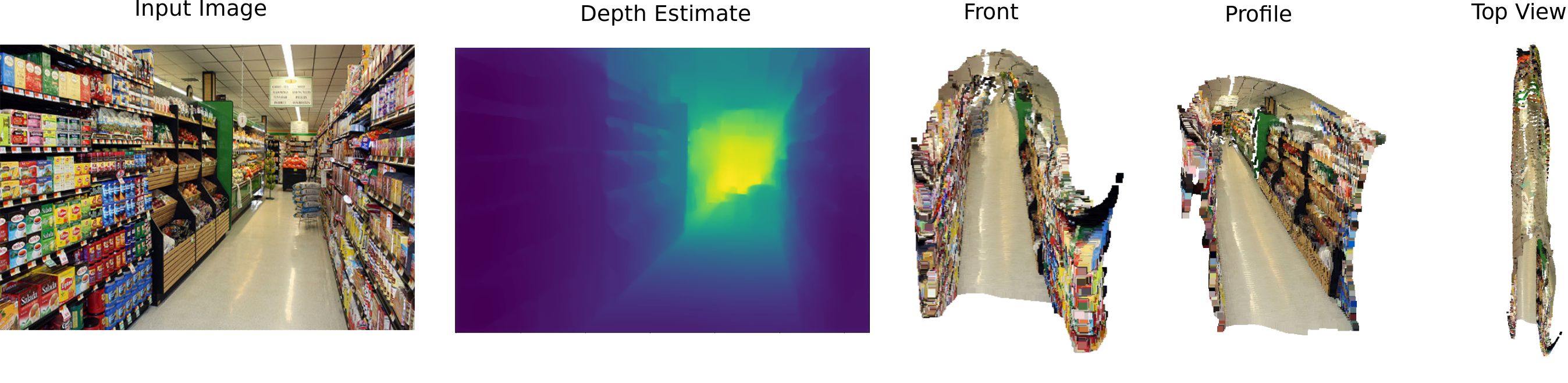}
    \caption{Monocular image to 3D point cloud processing pipeline. The original 2D image is used to infer the depth at different distances. In addition, a height map defining the distance of each pixel from the floor is generated. Finally, we synthesize a 3D point cloud of the environment.}
    \label{fig:pcl}
\end{figure*}

We argue that a symbiotic integration of physical sensing and ML-based inference enables PO\&G robots that inhabit the sweet spot between high cost-efficiency and high sensing capabilities. To this end, we show early results that integrate information from monocular depth estimation with positional information gathered via Bluetooth low-energy (BLE). The overall objective is to enable a retail robot, see Fig.~\ref{fig:robot}, to navigate in a retail environment and grasp and manipulate objects. To that end, 3D information about its position, the environment, and important objects needs to be available. As input sensors, we will only rely on a monocular camera (no depth information), BLE sensors, and the internal robot IMU. All of these sensors together can be acquired today for less than $\$10$ and typically weigh less than 200 grams.                

\section{System Architecture for 3D Mapping and Planogram Generation }
The robot platform used in this project consists of three main components: a robotic mobile base, in this case UGV jackal from clearpath robotics, a 6-DOF manipulator, and peripheral sensors, see Fig.~\ref{fig:robot}. In terms of processing power, the Jackal base is equipped with core-i5 dual-core processors, 8 GB RAM, and 32 GB ROM. Also, the Mobile base has inbuilt IMU and wheel encoders which provides odometry information. 

\begin{figure}[t!]
    \centering
    \includegraphics[width=\textwidth]{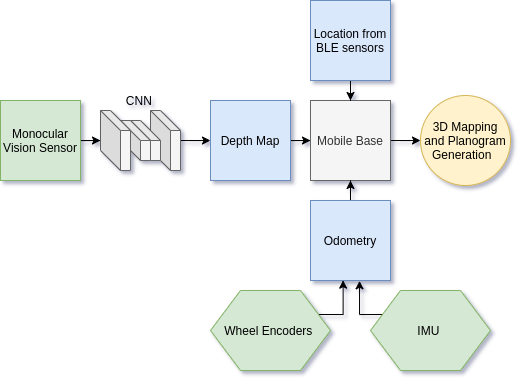}
    \caption{System architecture.}
    \label{fig:sysarch}
\end{figure}

The system processes three different sources of information namely information from a monocular camera, information from odometry, and information coming from external BLE sensors. Our overall objective is to combine the information from all of these sources to achieve high-quality 3D maps of the environment while also localizing the robot. Data from all three modalities are combined in order to maximize knowledge acquisition while also incorporating redundancies. The overall system architecture can be seen in Fig.~\ref{fig:sysarch}. Subsequently, we will discuss each component in more detail.

The first component in our system is a depth prediction network that generates per-pixel depth estimates of the environment. Our work builds upon recent work in computer vision \cite{alhashim2018high, wofk2019fastdepth,tateno2017cnn,laina2016deeper}. Our specific neural network architecture consists of an encoder-decoder network with skip connections between the encoder and decoder layers. Bilinear upsampling is used in the decoder to generate increasingly higher-resolution versions of the depth estimate. The loss function used in training combines the pixel wise Euclidean norm between predicted and ground truth depth maps, the Euclidean norm of the gradient of predicted and ground truth depth maps, and finally the structural similarity loss between the two images. The overall accuracy of our network on test data is approx. $91\%$ . 

\begin{figure*}[t!]
    \centering
    \includegraphics[width=18cm, height=4.5cm]{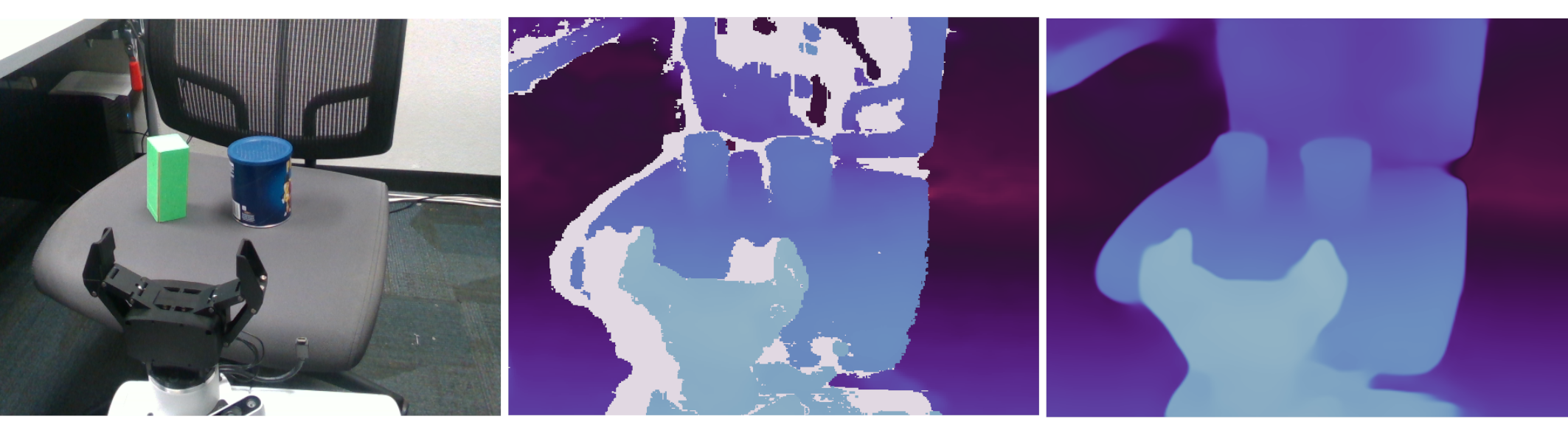}
    \caption{Output depth map (right) from modified CNN given input image (left) and ground truth depth (middle) obtained for test data (\href{https://www.dropbox.com/s/zn8wv420sy7bzy3/demo.mp4?dl=0}{demo})}
    \label{fig:results}
\end{figure*}

Once a depth map is generated, we use it to produce additional data structures. In particular, we generate a heightmap which stores the distance of each pixel to the floor. To this end, we perform plane fitting and estimate the normal of the largest plane at the bottom of the image. Given the RGB image and depth image, we also generate the 3D point cloud corresponding to the environment. In contrast to other low-cost approaches to depth estimation, e.g., Kinect, the approach presented can generate depth values for objects that are further away. Fig.~\ref{fig:pcl} shows an example of the processing pipeline, along with 3D views of the scene generated from a single input image. We can see, in particular in the 3D top view, that the corridor was fairly accurately approximated from a single image. Fig. \ref{fig:results} shows additional results of depth prediction vs ground truth in the form of point clouds. We notice that typical holes found in depth sensing devices are not present in the trained models. In addition, the approach also works for objects that are very close to the camera ($<20$cm) and, hence, is particularly well-suited for grasping and manipulation as seen in Fig.~\ref{fig:results}. Inference in our model and point cloud generation is performed at a rate of 9Hz.

In order to combine multiple point clouds into a map, traditional localization and mapping techniques and iterative closest-point methods can be used. Unfortunately, this often comes with substantial parameter tweaking and calibration which is a hindrance to PO\&G systems. We overcome this challenge by leveraging BLE sensors \cite{ozer2016improving, jianyong2014rssi}. In particular, we placed around 20 BLE sensors randomly in the workspace of the robot. These sensors can be acquired at extremely low-costs and do not require any calibration. The relative distance to each beacon can be deduced from the signal strength. A more complex operation can be used to calculate the relative orientation of the robot wrt. each beacon. Both sensor readings are, in turn, are used within an Extended Kalman filter to estimate the position and orientation of the robot in space. These positions are used as initial locations for the individual point clouds generated in the process described above, which are then refined through optimization techniques. Hence, larger maps of the environment can be obtained by leveraging both sources of information. IMU and wheel encoder data is also used within the localization process. Data from the BLE sensors is obtained and updated at a rate of 10 Hz. 

\section{Discussion}
In this paper, we discussed ongoing work regarding the development of a power-on-and-go robot system for retail environments. Eschewing the need for expensive sensing devices, we leverage multimodal integration and data processing in order to produce high-quality data from low-cost/low-quality sensors. The presented work is only a very first step towards a general principle for affordable robot systems that leverage machine learning abilities in order to fuse data from multiple, noisy sources into a rich data representation of the environment. Interestingly, our early results indicate that the trained systems may even be able to overcome the shortcomings and limitations of higher-end sensors. More specifically, the trained depth and point cloud estimation network was able to generate accurate maps of nearby objects, even in close distance situation and without any missing values. We hypothesize that this feature is due to the ability of neural networks to interpolate data. At the moment, the multimodal integration of the data is largely performed through hand-coded operations and procedures. As a next step, we are envisioning to learn the data fusion step itself. In doing so, the hope is to discover complex methods for data fusion that can cross-reference information across multiple modalities in order to yield a single, rich model of the world.

\section*{Acknowledgments}
This work is funded through research grant provided by Intel Corporation. 
%% Use plainnat to work nicely with natbib. 
\bibliographystyle{ieeetr}
\bibliography{references}

\end{document}